\documentclass{article}

\PassOptionsToPackage{numbers,sort&compress}{natbib}
\usepackage{genai4health_2026}

\usepackage[utf8]{inputenc} % allow utf-8 input
\usepackage[T1]{fontenc}    % use 8-bit T1 fonts
\usepackage{hyperref}       % hyperlinks
\usepackage{url}            % simple URL typesetting
\usepackage{booktabs}       % professional-quality tables
\usepackage{amsfonts}       % blackboard math symbols
\usepackage{nicefrac}       % compact symbols for 1/2, etc.
\usepackage{microtype}      % microtypography
\usepackage{xcolor}         % colors
\usepackage{amsmath}
\usepackage{amssymb} % for \checkmark
\usepackage{float}
\usepackage{longtable}
\usepackage{fvextra}
\usepackage{graphicx}
\usepackage{placeins}
\DefineVerbatimEnvironment{Prompt}{Verbatim}{
  fontsize=\scriptsize,
  breaklines=true,
  breakanywhere=true
}

\title{PathGeneLink: Constructing Evidence-Grounded Gene–Histomorphology Associations toward Morphology-Grounded Evaluation of Virtual Molecular Predictions}

\author{%
  \textbf{Arjun Beniwal}\textsuperscript{1,2} \quad
  \textbf{David R. Chang}\textsuperscript{1} \quad
  \textbf{Cheng-Han Wu}\textsuperscript{1} \\
  \textbf{Xuan Gong}\textsuperscript{1} \quad
  \textbf{Kun-Hsing Yu}\textsuperscript{1} \\
  \textsuperscript{1}Harvard Medical School, Harvard University, Boston, MA, USA \\
  \textsuperscript{2}New York University, New York, NY, USA
}

\begin{document}

\maketitle

\begin{abstract}
Virtual spatial-transcriptomics models that infer gene expression from H\&E are
typically evaluated using gene-wise correlation and related measures comparing
predicted with measured expression. These metrics quantify molecular agreement,
but they do not test whether a predicted molecular state is consistent with the
H\&E-visible morphology expected from prior biological knowledge. We introduce
\textbf{PathGeneLink}, an evidence-grounded human--AI pipeline for constructing
directional gene-expression-to-histomorphology relationships for
morphology-grounded evaluation. PathGeneLink first generates candidate
relationships between gene functional classes, expression states, and named
histomorphology, grounds them in biomedical literature, verifies the supporting
evidence, and subjects them to clinician review. Of 333 assumptions reviewed by
clinicians across 33 morphology features, 256 were retained. PathGeneLink then
links these reviewed assumptions to individual genes by checking whether the
gene's own evidence supports the biological function described in the
assumption. We validate this gene-specific linking step against a blinded
expert-clinician review of 379 gene--assumption pairs spanning 12 genes.
PathGeneLink reached \textbf{88.9\% agreement} with the clinician, with
\textbf{precision 0.77} and \textbf{recall 0.83} on the evaluated pairs
(Cohen's $\kappa=0.72$, 95\% CI 0.64--0.80). PathGeneLink achieved the
highest precision, recall, and $\kappa$ among the compared LLM conditions,
with the largest improvement when the reviewed biological assumption was
provided and a further numerical gain from gene evidence. These results support
PathGeneLink as a scalable
pipeline for constructing evidence-grounded gene--histomorphology knowledge,
with the resulting knowledge intended for future morphology-grounded evaluation
of virtual molecular predictions.
\end{abstract}

% Place Figure 1 at the top of page 2 (early top-only float).

\begin{figure}[!t]
    \centering
    \includegraphics[width=\textwidth]{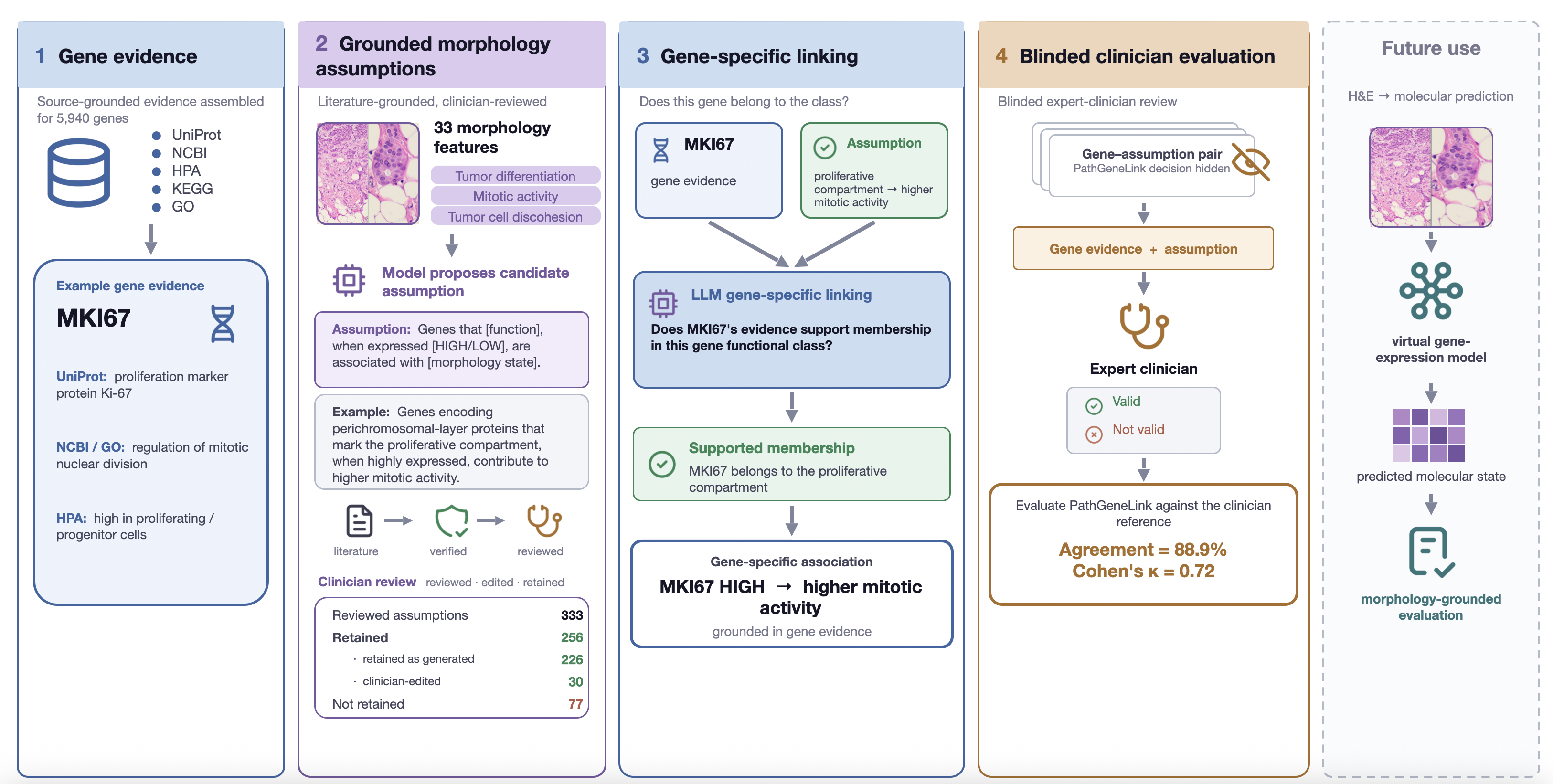}
    \caption{\textbf{Overview of the PathGeneLink workflow.}
    Gene evidence and literature-grounded, clinician-reviewed morphology
    assumptions are combined to make gene-specific functional-class membership
    decisions. A blinded expert-clinician review provides the reference used to evaluate these associations, with future application to
    morphology-grounded evaluation of H\&E-to-molecular prediction models.}
    \label{fig:pathgenelink-workflow}
\end{figure}

\section{Introduction}

Spatial transcriptomics maps gene expression back onto intact tissue, making it
possible to study how molecular programs vary across regions, cell populations,
and the tumor microenvironment. This is especially useful in cancer, where
spatial relationships between tumor, immune, and stromal compartments can be
biologically and clinically important. But spatial transcriptomics remains
expensive and is not routinely available, motivating deep learning models that
infer spatial gene expression directly from standard H\&E slides. Recent
datasets and benchmarks have made this an increasingly active problem across
tissues, cancer types, and spatial transcriptomic platforms
\citep{jaume2024hest1k,wang2025benchmarking}.

Most evaluations ask whether predicted expression matches measured expression,
usually through gene-wise correlations. These measurements remain essential,
but they do not directly ask whether a predicted molecular state is consistent
with the biology visible in the tissue. Recent work has begun moving in this
direction --- evaluating predicted expression within molecular pathways or
against independently annotated tissue phenotypes
\citep{hu2026histoprism,nonchev2026deepspotm}, and relating molecular states to
interpretable histologic morphology from human-defined features, paired spatial
data, or learned pathology representations
\citep{diao2021humaninterpretable,bae2021spade,huang2025morphlink,kim2026picasso,muench2026concept}
(Section~\ref{sec:related}).
Together, these efforts motivate a complementary question for evaluating
virtual molecular predictions: \textbf{if a model predicts a molecular state in a
tissue region, is the H\&E morphology consistent with what biology would lead
us to expect?}

Answering this systematically requires structured knowledge of what morphology is expected for a given molecular state. Existing resources describe gene
function, expression, pathways, disease associations, and phenotypes
\citep{uhlen2015humanproteome,kohler2021hpo,uniprot2025,kanehisa2025kegg,geneontology2026,cerami2012cbioportal},
but to our knowledge there is no structured knowledge resource that directly captures
directional relationships of the form
\emph{gene expression HIGH/LOW $\rightarrow$ named H\&E-visible morphology},
together with the evidence supporting that relationship. Such knowledge would
enable a \textbf{morphology-grounded evaluation} complementary to expression
correlation: a model could be evaluated not only on whether its predicted
expression matches measured expression, but also on whether its predictions
agree with expected tissue morphology. 

We introduce \textbf{PathGeneLink}, a human--AI pipeline for constructing
these directional relationships at scale. Rather than directly asking an LLM
whether a particular gene should be associated with a particular morphology,
PathGeneLink decomposes the gene-specific association into two components:
evidence-supported membership of a gene in a functional class, and a
clinician-reviewed directional association between that functional class at a
specified expression state and a morphology state:

\[
\underbrace{\text{gene} \in \text{gene functional class}}
_{\text{evidence-supported membership}}
\;+\;
\underbrace{(\text{gene functional class},\ \text{expression state})
\rightarrow \text{morphology state}}
_{\text{reviewed directional association}}.
\]

Gene-centered evidence is assembled from curated biomedical resources.
Candidate relationships between gene functional classes, expression direction,
and morphology are grounded in the biomedical literature, their supporting
citations and evidence are verified, and clinicians then retain, revise, or
reject them. Only clinician-reviewed assumptions are used for gene-specific
linking. PathGeneLink then applies these general relationships across
individual genes by asking a narrower question: \textbf{does the evidence for
this gene support membership in the functional class described by the
assumption?} Figure~\ref{fig:pathgenelink-workflow} summarizes the complete
workflow and its intended use for morphology-grounded evaluation.

This separation keeps the general function-to-morphology relationship grounded
and clinician reviewed while allowing the gene-specific membership step to be
applied at scale. We evaluate this automated linking step against a blinded
expert-clinician review. The present work therefore focuses on constructing the biological
knowledge resource and evaluating the reliability of its scalable gene-specific
expansion; applying the resulting associations to H\&E-to-molecular prediction
models is the next step.

Our contributions are twofold:
\begin{enumerate}
    \item \textbf{PathGeneLink}, an evidence-grounded human--AI pipeline for
    constructing directional gene-expression-to-histomorphology relationships
    at scale, using literature-grounded, clinician-reviewed biological
    assumptions and evidence-grounded gene-specific linking, intended for
    \textbf{morphology-grounded evaluation of H\&E-to-molecular prediction
    models}.

    \item \textbf{A blinded expert-clinician evaluation of the
    scalable gene-specific linking step across 12 genes}, directly measuring how
    reliably the automated system constructs these associations.
\end{enumerate}

\section{Related work}
\label{sec:related}

\paragraph{Biological evaluation of virtual molecular predictions.}
Histology-to-expression models are commonly evaluated by comparing predicted
and measured expression, with \textbf{gene-wise correlation} among the most
widely reported measures \citep{jaume2024hest1k}. Yet correlation alone can mask
biologically meaningful performance --- it is inflated by the large fraction of
low- or non-expressed genes --- motivating complementary metrics for
generalizability and translational potential \citep{wang2025benchmarking}.
Recent work has begun asking more biologically structured questions. HistoPrism groups genes into known Hallmark
and Gene Ontology pathways and evaluates how accurately expression is recovered
within each pathway \citep{hu2026histoprism}. Its evaluation therefore still
compares \textbf{predicted and measured molecular expression}, but asks whether
coherent biological programs are recovered rather than considering only
individual highly variable genes. DeepSpot-M demonstrates a different kind of
validation: where reliable molecular ground truth was unavailable, predicted
TLS-marker expression was evaluated against expert-annotated tertiary lymphoid
structures in the tissue \citep{nonchev2026deepspotm}. This provides an
important precedent for using morphology as an \textbf{orthogonal evaluation
signal} for virtual molecular predictions.

PathGeneLink aims to make this kind of morphology-grounded evaluation
systematic. Rather than defining a study-specific pathology check for one
molecular program, it constructs reusable directional expectations linking
molecular states to named H\&E phenotypes. Because these relationships are
constructed independently of the model being evaluated, they can provide an external knowledge resource for evaluation rather than being inferred from the same predictions they
are used to assess.

\paragraph{Morphology--molecular relationships.}
The relationship between tissue morphology and molecular state is itself well
established. Diao et al.\ derived human-interpretable image features from
cell- and tissue-classification models trained on pathologist annotations and
used them to predict diverse
molecular phenotypes \citep{diao2021humaninterpretable}. SPADE integrates
co-registered spatial transcriptomics and histology to identify genes associated
with learned morphological image features \citep{bae2021spade}.
MorphLink extracts interpretable quantitative morphology features from
histology and uses paired spatial-omics measurements to identify
morphology--molecular features with similar spatial patterns
\citep{huang2025morphlink}. More recently, PICASSO discovers
human-interpretable histomorphologic concepts from pathology foundation-model
representations and relates them to spatial gene-expression programs
\citep{kim2026picasso}, while Muench et al.\ identify morphological concepts
that explain virtual gene-expression predictions from H\&E
\citep{muench2026concept}.

These approaches demonstrate that informative morphology--molecular
relationships can be \textbf{discovered from paired spatial data or from the
models being interpreted}. PathGeneLink addresses a different question:
\textbf{can expected relationships be specified independently of both the paired
evaluation cohort and the predictor being evaluated?} It specifies such
expectations directly, from literature-grounded, clinician-reviewed assumptions.

\paragraph{Grounded biological knowledge construction.}
Existing biomedical resources provide much of the evidence needed to reason
about genes but do not directly provide this evaluation object. The Human
Protein Atlas provides tissue- and cancer-level RNA and protein-expression
resources
\citep{uhlen2015humanproteome,uhlen2017pathologyatlas}, while the Human
Phenotype Ontology provides a standardized ontology of human phenotypic
abnormalities and disease annotations \citep{kohler2021hpo}. Neither directly
represents directional gene-expression-state$\rightarrow$H\&E-morphology
relationships suitable for evaluating virtual molecular predictions.

LLMs can help synthesize biological knowledge distributed across such resources
and the literature. Hu et al.\ showed that LLMs can recover coherent functions
for gene sets and provide rationales that are often expert-verifiable
\citep{hu2025geneset}, while GeneAgent showed that checking LLM-generated
biological claims against domain databases improves their reliability
\citep{wang2025geneagent}. PathGeneLink uses LLMs for scale, but keeps the
biological knowledge anchored to source evidence: gene reasoning starts from
curated gene databases, morphology assumptions are grounded in retrieved
biomedical literature and independently evidence-verified, and the assumptions
used downstream are directly reviewed by clinicians. The goal is not automated
knowledge generation itself, but a scalable way to construct a sufficiently reliable \textbf{knowledge resource} linking molecular state to interpretable tissue morphology.

%%%%%%%%%%%%%%%%%%%%%%%%%%%%%%%%%%%%%%%%%%%%%%%%%%%%%%%%%%%%

\section{Methods}

PathGeneLink constructs directional gene--histomorphology associations by
separating two questions: \textbf{what morphology is expected from a biological
function}, and \textbf{which individual genes perform that function}. We first
assemble source-grounded evidence for individual genes. Separately, we construct
and clinically review general relationships between gene function, expression
state, and histomorphology. We then apply these reviewed relationships to
individual genes and evaluate the resulting gene-specific associations against
blinded clinician judgments.

\subsection{Gene evidence resource}

We assembled evidence for individual genes from public biomedical resources,
including UniProt \citep{uniprot2025}, the National Center for Biotechnology
Information (NCBI) \citep{sayers2025ncbi}, the Human Protein Atlas
\citep{uhlen2015humanproteome,uhlen2017pathologyatlas}, the Kyoto Encyclopedia
of Genes and Genomes (KEGG) \citep{kanehisa2025kegg}, Gene Ontology (GO)
\citep{geneontology2026}, and cBioPortal \citep{cerami2012cbioportal}. Gene
symbols and aliases were resolved before retrieval. The resulting resource
contains \textbf{5,940 genes with retrieved evidence}.

The collected evidence covers gene function, biological processes and pathways,
expression, cellular localization, and disease context. Source text is retained
with its provenance rather than replaced by an LLM-generated summary.
Source-specific collection procedures are described in
Appendix~\ref{app:gene-evidence}.

\subsection{Grounded and clinician-reviewed morphology assumptions}

We consider \textbf{33 histomorphology features}, represented through
biologically meaningful states such as \emph{poorly differentiated},
\emph{higher mitotic activity}, or \emph{perineural invasion present}. The complete morphology feature vocabulary and its grading states are listed in
Appendix~\ref{app:morphology-features}. We
define an \textbf{assumption} as a natural-language relationship of the form:

\begin{quote}
\textbf{``Genes that [BIOLOGICAL FUNCTION], when expressed at [HIGH/LOW]
levels, are associated with [MORPHOLOGY STATE].''}
\end{quote}

The biological function defines a \textbf{gene functional class}, while the
expression and morphology states specify the directional relationship. For the spatial-transcriptomics setting considered here, expression state
refers to RNA abundance.

A large language model (LLM) proposes candidate causal and correlative
assumptions. Supporting biomedical literature is then retrieved, and the cited
papers and quoted evidence are verified against their sources. The evidence for
each candidate is assessed and assigned one of four tiers:
\textbf{established}, for direct support; \textbf{supported}, when the
relationship is supported through an intermediate biological step or contextual
qualification; \textbf{weak}, for partial support; and \textbf{ungrounded},
when the retrieved evidence does not support the proposed relationship. Only
\emph{established} and \emph{supported} candidates are eligible for inclusion
in the reviewed resource. Detailed tier definitions and evidence-verification procedures are provided in
Appendix~\ref{app:assumptions}, with the models used and full prompts reported in
Appendix~\ref{app:prompts}.

Two physician reviewers divided the
morphology features between them. For each candidate, reviewers were shown the
proposed assumption together with its supporting papers and quoted evidence and
could \textbf{retain, revise, or reject} it. Only assumptions retained after
clinician review are used for gene-specific linking.

\subsection{Applying reviewed assumptions to individual genes}

The reviewed assumptions describe how a \textbf{biological function} at a
specified expression state relates to a morphology state, but do not specify
every individual gene that performs that function. PathGeneLink therefore
factorizes a gene-specific association into

\[
\underbrace{\text{gene} \in \text{gene functional class}}
_{\text{automated membership decision}}
\;+\;
\underbrace{(\text{gene functional class},\ \text{expression state})
\rightarrow \text{morphology state}}
_{\text{reviewed assumption}}.
\]

For each gene and reviewed assumption, the model receives the gene evidence
from Section~3.1 and asks whether that evidence supports membership of the gene
in the functional class described by the assumption. The reviewed assumption
already specifies the corresponding expression direction and morphology state;
the gene-specific model therefore does not infer a new morphology relationship
at this stage.

If gene membership is supported, PathGeneLink instantiates the reviewed
assumption for that gene, producing a directional gene-specific association.
Thus, the automated decision is the \textbf{gene-to-functional-class} link,
while the downstream expression and morphology states are inherited from the
reviewed assumption. Membership must be supported by the corresponding gene
evidence, and source provenance is retained with the resulting association.
Contextual qualifications in the reviewed assumption, including organ scope
where applicable, are retained in the gene-specific association.
Additional model, prompting, verification, and inference details are provided
in Appendix~\ref{app:gene-linking}. Abridged end-to-end examples showing the
gene evidence, reviewed assumption, gene-specific decision, and clinician
reference label are provided in Appendix~\ref{app:end-to-end-examples}.

\subsection{Blinded clinician evaluation}

We evaluated the gene-specific linking step on \textbf{12 genes} (CDH1, MKI67,
ACTA2, CD8A, EGFR, PTEN, CTNNB1, BRCA1, PECAM1, MS4A1, CD68, and VIM), spanning
epithelial, proliferative, stromal, immune, vascular, and DNA-repair biology.

The evaluation set contains \textbf{379 gene--assumption pairs}. It
includes associations emitted by PathGeneLink together with non-emitted
candidates selected either by embedding similarity between the gene evidence
and the corresponding functional-class description or at random. The
similarity ranking uses the gene evidence and functional-class text rather than
PathGeneLink scores or confidence, providing a set of topically similar
non-emitted candidates alongside a broader random sample. The complete sampling
procedure and construction of the evaluation set are described in
Appendix~\ref{app:evaluation-set}. The set includes all emitted associations
for these genes and sampled non-emitted candidates; the reported metrics
reflect this evaluation set.

A single expert clinician reviewed each pair while blinded to the
PathGeneLink decision and to the sampling source of each pair. For each pair,
the clinician was shown the gene evidence and reviewed assumption and judged
whether the gene belonged to the functional class described by that assumption
-- the same gene-membership question posed by the automated linking step. These
clinician judgments provide the reference labels used for evaluation.

We report raw agreement, precision, recall, and Cohen's $\kappa$
\citep{cohen1960kappa}, together with 95\% bootstrap confidence intervals;
precision and recall use the clinician labels as reference. Metric definitions
and confidence-interval estimation are provided in
Appendix~\ref{app:statistics}.

Finally, we compare PathGeneLink with three matched LLM baselines on the same
evaluation set: \textbf{Direct LLM}, \textbf{LLM + gene evidence}, and
\textbf{LLM + reviewed assumption}. The target gene--assumption pairs and clinician reference labels are held
fixed across conditions, while the available gene evidence and reviewed
assumption differ. The Direct LLM and gene-evidence-only conditions assess the
gene--morphology association without the functional-class description supplied
to the assumption-based conditions.

%%%%%%%%%%%%%%%%%%%%%%%%%%%%%%%%%%%%%%%%%%%%%%%%%%%%%%%%%%%%

\section{Results}

\subsection{Assumption curation}

Clinicians reviewed \textbf{333 candidate assumptions} across the 33 morphology
features. Of these, \textbf{256 were retained in the final resource}: 226
without revision and 30 after clinician revision. The remaining 77 were not
retained. All assumptions in the final resource satisfied the
\emph{established (103)} or \emph{supported (153)} evidence criteria defined
in Section~3.2. Table~\ref{tab:assumption-curation} summarizes the curation
outcomes.

\begin{table}[!htbp]
\caption{Assumption curation.}
\label{tab:assumption-curation}
\centering
\small
\begin{tabular*}{0.62\linewidth}{@{\extracolsep{\fill}}lr@{}}
\toprule
\textbf{Measure} & \textbf{n} \\
\midrule
Assumptions reviewed & \textbf{333} \\
Retained in final resource & \textbf{256} \\
\quad Retained without revision & 226 \\
\quad Retained after clinician revision & 30 \\
Not retained & \textbf{77} \\
\bottomrule
\end{tabular*}
\end{table}

\FloatBarrier

\subsection{PathGeneLink agreement with the clinician reference}

A single expert clinician reviewed all \textbf{379 gene--assumption pairs},
marking \textbf{102 as Valid} and 277 as Not valid; these labels form the
evaluation reference. Against them, PathGeneLink reached \textbf{88.9\%
agreement} with the clinician (95\% CI 86--92) and \textbf{Cohen's
$\kappa=0.72$} (95\% CI 0.64--0.80). Of the 110 associations emitted by
PathGeneLink, 85 were Valid in the reference, corresponding to a
\textbf{precision of 0.77}; PathGeneLink identified 85 of the 102 Valid
associations, a \textbf{recall of 0.83}.

We then compared PathGeneLink with three matched LLM conditions on the same
evaluated pairs (Table~\ref{tab:baseline-comparison}).
These conditions were designed to isolate the contribution of retrieved gene
evidence and the clinician-reviewed assumption.

\begin{table}[!htbp]
\caption{Comparison with matched LLM baselines on the 379 evaluated pairs, with the clinician labels as reference. Values in brackets are 95\% confidence intervals.}
\label{tab:baseline-comparison}
\centering
\small
\setlength{\tabcolsep}{2.7pt}
\renewcommand{\arraystretch}{1.08}
\begin{tabular}{@{}lccccc@{}}
\toprule
\textbf{Method}
& \textbf{Gene}
& \textbf{Reviewed}
& \textbf{Precision}
& \textbf{Recall}
& \textbf{Cohen's $\boldsymbol{\kappa}$} \\
& \textbf{evidence}
& \textbf{assumption}
& & & \\
\midrule

Direct LLM
& -- & --
& 0.49 [0.38, 0.60]
& 0.42 [0.32, 0.52]
& 0.27 [0.17, 0.38] \\

LLM + gene evidence
& \checkmark & --
& 0.51 [0.39, 0.62]
& 0.36 [0.27, 0.46]
& 0.26 [0.15, 0.36] \\

LLM + reviewed assumption
& -- & \checkmark
& 0.72 [0.63, 0.80]
& 0.73 [0.64, 0.82]
& 0.62 [0.53, 0.71] \\

\textbf{PathGeneLink}
& \checkmark & \checkmark
& \textbf{0.77 [0.69, 0.85]}
& \textbf{0.83 [0.76, 0.90]}
& \textbf{0.72 [0.64, 0.80]} \\

\bottomrule
\end{tabular}
\end{table}

\FloatBarrier

Providing the reviewed assumption improved agreement over the Direct LLM and
gene-evidence-only conditions. Adding gene evidence yielded a further numerical
gain, with PathGeneLink achieving the highest precision, recall, and $\kappa$
while grounding membership decisions in source evidence.

%%%%%%%%%%%%%%%%%%%%%%%%%%%%%%%%%%%%%%%%%%%%%%%%%%%%%%%%%%%%

\section{Discussion and Limitations}

PathGeneLink addresses a complementary aspect of virtual molecular-model
evaluation: whether a predicted molecular state is consistent with the tissue
morphology expected from prior biological knowledge. Rather than deriving this
expectation from the model being evaluated, PathGeneLink constructs it
independently by combining literature-grounded morphology assumptions,
clinician review, and gene-specific evidence. The resulting
gene-specific associations showed strong agreement with the expert clinician
(88.9\% agreement, $\kappa=0.72$), supporting the feasibility of constructing this
knowledge resource at scale.

The baseline comparison indicates the clinician-reviewed assumption as the main
source of this agreement, with retrieved gene evidence adding a further gain and
supplying the external grounding for each membership decision. How much the
evidence contributes for less-characterized genes, beyond the 12 evaluated here,
remains an open question.

The intended use of PathGeneLink is \textbf{alongside}, rather than instead of,
conventional molecular evaluation. Expression correlation measures whether
predicted expression matches measured expression; PathGeneLink is intended to
ask an additional question: whether those predictions are also consistent with
expected H\&E morphology. The next step is therefore to apply the knowledge resource to
real H\&E-to-spatial-transcriptomics models on paired tissue data and test
whether morphology-grounded evaluation provides information complementary to
expression-based metrics. More broadly, the same framework could support other
virtual molecular modalities where reliable directional
molecular--morphology relationships can be established.

Several limitations remain. The gene-specific evaluation spans 12 biologically
diverse genes reviewed by a single expert clinician. Expanding the
evaluation across more genes, morphology features, and additional clinical
experts, with broader sampling of non-emitted candidates, will be important
for establishing how consistently the approach generalizes. In addition,
this study evaluates the construction and reliability
of the PathGeneLink knowledge resource itself; its value for
morphology-grounded evaluation of virtual molecular predictions remains to be
tested directly on paired H\&E and spatial-omics data.

These results provide an initial validation of PathGeneLink as a scalable
approach for constructing directional gene--histomorphology knowledge grounded
in source evidence and clinician-reviewed biological assumptions, on which
morphology-grounded evaluation of virtual molecular predictions can be built.

\paragraph{Materials availability.}
The morphology feature vocabulary, full prompt templates, and illustrative
gene-specific associations are provided in the Appendix.

%%%%%%%%%%%%%%%%%%%%%%%%%%%%%%%%%%%%%%%%%%%%%%%%%%%%%%%%%%%%

% \section*{References}

\bibliographystyle{unsrtnat}
\bibliography{references}

%%%%%%%%%%%%%%%%%%%%%%%%%%%%%%%%%%%%%%%%%%%%%%%%%%%%%%%%%%%%

%%%%%%%%%%%%%%%%%%%%%%%%%%%%%%%%%%%%%%%%%%%%%%%%%%%%%%%%%%%%
\appendix

\section{Gene evidence and morphology feature vocabulary}
\label{app:input-resources}

\subsection{Gene evidence collection}
\label{app:gene-evidence}

PathGeneLink constructs a source-grounded evidence record for each gene before
gene-specific linking. Gene symbols and aliases are first resolved against the
NCBI human gene reference. Exact official-symbol matches take precedence over
synonym matches; unique deprecated synonyms are mapped to the corresponding
official gene, while ambiguous mappings are excluded from automated use.

The input panel contains 5,969 symbols. Of these, 5,947 resolve confidently to
human genes, and 5,940 have retrieved evidence used by PathGeneLink. Evidence
is collected from the sources summarized in
Table~\ref{tab:app-gene-sources}. Descriptive source text is retained with its
source and provenance rather than replaced by an LLM-generated summary.

\begin{table}[h]
\caption{Biomedical sources used to construct gene evidence records.}
\label{tab:app-gene-sources}
\centering
\small
\setlength{\tabcolsep}{5pt}
\begin{tabular}{@{}p{0.17\linewidth}p{0.55\linewidth}p{0.20\linewidth}@{}}
\toprule
\textbf{Source} & \textbf{Information retained} & \textbf{Representation} \\
\midrule

UniProt &
Gene function, tissue specificity, subcellular location, disease, and
protein-interaction information &
Source text + accession \\

NCBI Gene &
RefSeq gene summary and aliases &
Source text + Entrez ID \\

Human Protein Atlas &
RNA/protein expression, biological process, localization, tissue and cancer
context &
Source text / reported values + Ensembl ID \\

KEGG &
Pathways, diseases, and drug-target annotations &
Source annotations + Entrez ID \\

Gene Ontology &
Biological Process, Molecular Function, and Cellular Component annotations &
Structured GO categories \\

cBioPortal &
Cancer-genomic and expression summaries across TCGA PanCancer Atlas studies &
Computed summaries + study provenance \\

\bottomrule
\end{tabular}
\end{table}

Retrieval is identifier-driven where possible: NCBI and KEGG are queried by
Entrez identifier, the Human Protein Atlas by Ensembl identifier, and UniProt
through reviewed human entries. Retrieved evidence remains associated with its
original database section so that functional evidence can be distinguished
from expression, localization, or disease-context evidence. 

The evidence resource is intentionally heterogeneous: not every source provides
information for every gene. The gene-specific linking stage therefore reasons
only from the evidence actually retrieved for the gene rather than assuming
that absence from a particular source indicates absence of the corresponding
biology.

\subsection{Morphology feature vocabulary}
\label{app:morphology-features}

PathGeneLink uses 33 histomorphology features. Each feature is associated with
a set of grading states that defines the morphology axis used during assumption
generation. Table~\ref{tab:app-morphology-features} lists the complete feature
vocabulary used in the final resource.

\begin{longtable}{@{}p{0.43\linewidth}p{0.49\linewidth}@{}}
\caption{Morphology feature vocabulary used by PathGeneLink.}
\label{tab:app-morphology-features}\\

\toprule
\textbf{Morphology feature} & \textbf{Grading states} \\
\midrule
\endfirsthead

\toprule
\textbf{Morphology feature} & \textbf{Grading states} \\
\midrule
\endhead

\bottomrule
\endfoot

Signet ring cell morphology
& absent / low-focal / predominant \\

Cellular pleomorphism
& low-none / moderate / high-marked \\

Nuclear atypia
& low-absent / moderate / high-severe \\

Mucinous cytoplasm
& absent-low / prominent \\

Nuclear pleomorphism / nuclear uniformity
& uniform-low / moderate / marked \\

Chromatin condensation
& condensation / decondensation \\

Nucleolar prominence
& absent-inconspicuous / conspicuous-prominent \\

Nuclear-to-cytoplasmic ratio
& low / intermediate / high \\

Nuclear grade
& low / intermediate / high \\

Nuclear size / anisonucleosis
& normal-small / enlarged-variable \\

Tumor cellularity
& low-sparse / moderate / high-hypercellular \\

Mitotic activity / mitotic figures
& low-absent / intermediate / high \\

Cellular proliferative activity
& low / high \\

Apoptosis
& low-absent / high-prominent \\

Tumor necrosis
& low-absent / high-prominent \\

Histologic tumor grade
& low / intermediate / high \\

Degree of tumor differentiation
& well-differentiated / poorly differentiated \\

Squamous differentiation
& absent / present \\

Neuroendocrine differentiation
& absent / present \\

Tumor cell discohesion
& cohesive / mixed / discohesive \\

Tumor budding
& low / intermediate / high \\

Gland formation
& low-poor / moderate / high-well formed \\

Tumor border / growth pattern
& pushing-well defined / mixed / infiltrative \\

Blood vessel / vascular invasion
& absent / present \\

Tumor capsule invasion
& absent / present \\

Lymphatic vessel invasion
& absent / present \\

Perineural invasion
& absent / present \\

Stromal reaction
& minimal-mature / reactive \\

Angiogenesis / vascular proliferation
& low-absent / moderate / high-prominent \\

Tumor vascularity / vascular density
& low-sparse / moderate / high-rich \\

Tumor-infiltrating lymphocytes
& low-absent / moderate / high-brisk \\

Tertiary lymphoid structures
& absent / present \\

Neoplastic / malignant status
& non-neoplastic / benign-in situ / invasive malignant \\

\end{longtable}

\section{Assumption generation, grounding, and clinician curation}
\label{app:assumptions}

\subsection{Candidate generation and grounding}

For each of the 33 morphology features, PathGeneLink constructs gene-agnostic
candidate assumptions of the form

\[
(\text{gene functional class},\ \text{expression state})
\rightarrow
\text{morphology state}.
\]

The assumption-generation pipeline contains three stages:

\begin{enumerate}
    \item \textbf{Author:} an LLM proposes candidate causal and correlative
    relationships between biological functions, expression direction, and the
    morphology feature.

    \item \textbf{Retrieve:} biomedical literature is searched for evidence
    supporting each candidate relationship. Returned PubMed identifiers are
    checked to ensure that they resolve, and quoted evidence is verified
    against the corresponding source abstract.

    \item \textbf{Verify:} a separate model evaluates the proposed relationship
    against the retrieved evidence and assigns an evidence tier.
\end{enumerate}

The generation and grounding procedure produced 551 candidate assumptions
across the 33 morphology features. Their evidence-tier distribution is shown
in Table~\ref{tab:app-evidence-tiers}.

\begin{table}[h]
\caption{Evidence tiers assigned during assumption grounding.}
\label{tab:app-evidence-tiers}
\centering
\small
\begin{tabular}{@{}lr@{}}
\toprule
\textbf{Evidence tier} & \textbf{n} \\
\midrule
Established & 125 \\
Supported & 260 \\
Weak & 122 \\
Ungrounded & 44 \\
\midrule
Total & \textbf{551} \\
\bottomrule
\end{tabular}
\end{table}

\subsection{Evidence-tier definitions}

The four evidence tiers are defined as follows:

\begin{itemize}
    \item \textbf{Established:} retrieved evidence directly supports the
    functional-class, expression-direction, and morphology relationship.

    \item \textbf{Supported:} the relationship is supported through an
    intermediate biological step, contextual qualification, or a clearly
    connected multi-source evidence chain.

    \item \textbf{Weak:} the retrieved evidence supports only part of the
    proposed relationship or is merely topically related.

    \item \textbf{Ungrounded:} the retrieved evidence does not support the
    proposed relationship, or supports a different biological intervention
    such as mutation or deletion rather than the stated expression-level
    relationship.
\end{itemize}

The \emph{ungrounded} label denotes lack of support in the retrieved evidence;
it is not interpreted as evidence that the biological relationship is absent.
Only \emph{established} and \emph{supported} assumptions are eligible for the
clinician-reviewed resource.

\subsection{Clinician curation}

Two physician reviewers divided the morphology features between them and
reviewed candidate assumptions together with their supporting papers and quoted
evidence. Reviewers could retain, revise, or reject each candidate.

Across the review process, after relevance- and diversity-based filtering, 333 distinct assumptions were selected and received clinician review.
The final resource contains 256 assumptions: 226 retained without revision and
30 retained after clinician revision. Seventy-seven reviewed candidates were
not retained. Among the final 256 assumptions, 103 satisfy the
\emph{established} evidence criterion and 153 satisfy the
\emph{supported} criterion. These counts correspond to the curation outcomes
reported in Table~\ref{tab:assumption-curation} in the main text.

\section{Gene-specific linking}
\label{app:gene-linking}

The gene-specific stage instantiates a reviewed assumption for an individual
gene. Its task is deliberately narrower than directly predicting a morphology
from a gene.

For each gene--assumption pair, the model receives:

\begin{itemize}
    \item the gene's source-grounded evidence record;
    \item the gene functional class specified by the reviewed assumption;
    \item the expression state and morphology state already specified by that
    assumption.
\end{itemize}

The model then answers only whether the supplied gene evidence supports
membership of that gene in the assumption's functional class. It does not
re-evaluate the function-to-morphology relationship or infer a new morphology
direction. Thus, the decision decomposes as

\[
\underbrace{\text{gene} \in \text{gene functional class}}
_{\text{automated membership decision}}
\;+\;
\underbrace{(\text{gene functional class},\ \text{expression state})
\rightarrow \text{morphology state}}
_{\text{inherited from reviewed assumption}}.
\]

Membership decisions must be supported by citations to the supplied gene
evidence. A candidate is emitted as a gene-specific association only when the
gene is classified as a member of the functional class and at least one valid
gene-evidence citation survives deterministic verification.

The primary PathGeneLink configuration uses a generation pass followed by an
adversarial verification pass. The first pass evaluates functional-class
membership from the supplied gene evidence. The verification  acts as an entailment layer and receives the
same evidence and assumptions and independently attempts to refute positive
membership decisions. Associations that fail the second check are removed.

The generation pass uses \texttt{gpt-5.4-mini}; the verification pass uses
\texttt{gpt-5.5}. Both operate under the same membership criterion. Full prompt
text is provided in Appendix~\ref{app:prompts}.

\subsection{Abridged end-to-end examples}
\label{app:end-to-end-examples}

Because complete gene-evidence records can be lengthy, we show abridged
examples illustrating how source-grounded gene evidence is combined with a
reviewed morphology assumption to produce a gene-specific association. All
four examples below are included in the clinician evaluation set,
were emitted by PathGeneLink, and have a clinician reference label of
\emph{Valid}. Evidence excerpts are taken from the source evidence cited by
the gene-specific linking model, and the morphology assumptions are the
reviewed assumptions used by PathGeneLink.

\begin{longtable}{@{}p{0.22\linewidth}p{0.72\linewidth}@{}}
\caption{Abridged end-to-end PathGeneLink examples from the clinician
evaluation set.}
\label{tab:app-end-to-end-examples}\\

\toprule
\textbf{Stage} & \textbf{Example} \\
\midrule
\endfirsthead

\toprule
\textbf{Stage} & \textbf{Example} \\
\midrule
\endhead

\bottomrule
\endfoot

\multicolumn{2}{@{}l}{\textbf{MKI67 -- Mitotic activity}} \\
\addlinespace[0.25em]

Gene evidence &
UniProt (FUNCTION): ``relocalizes \ldots{} to the chromosome surface where it
forms extended brush structures''; NCBI / GO: ``regulation of mitotic nuclear
division.'' \\

Reviewed morphology assumption &
Genes encoding cycling-cell nuclear-scaffold / perichromosomal-layer proteins
that mark the proliferative compartment, when expressed at \textbf{HIGH}
levels, are associated with \textbf{higher mitotic activity}. \\

PathGeneLink output &
\textbf{Member}; instantiated association:
\textbf{MKI67 HIGH $\rightarrow$ higher mitotic activity}. \\

Clinician evaluation &
Final clinician reference
label: \textbf{Valid}. \\

\midrule
\multicolumn{2}{@{}l}{\textbf{CDH1 -- Tumor cell discohesion}} \\
\addlinespace[0.25em]

Gene evidence &
UniProt: ``calcium-dependent cell adhesion proteins'';
``involved in \ldots{} cell-cell adhesions'';
``desmosome cell-cell junction formation.'' \\

Reviewed morphology assumption &
Adhesion-junction builders, including adherens-junction and desmosomal
functions, when expressed at \textbf{LOW} levels, are associated with
discohesive growth. \\

PathGeneLink output &
\textbf{Member}; instantiated association:
\textbf{CDH1 LOW $\rightarrow$ tumor cell discohesion}. \\

Clinician evaluation &
Final clinician reference label: \textbf{Valid}. \\

\midrule
\multicolumn{2}{@{}l}{\textbf{ACTA2 -- Cellular pleomorphism}} \\
\addlinespace[0.25em]

Gene evidence &
UniProt: ``involved in various types of cell motility'';
NCBI: ``smooth muscle actin \ldots{} involved in vascular contractility.'' \\

Reviewed morphology assumption &
Actin-remodeling, contractility, and motile shape-change functions, when
expressed at \textbf{HIGH} levels, are associated with
\textbf{marked cellular pleomorphism}. \\

PathGeneLink output &
\textbf{Member}; instantiated association:
\textbf{ACTA2 HIGH $\rightarrow$ marked cellular pleomorphism}. \\

Clinician evaluation &
Final clinician reference label: \textbf{Valid}. \\

\midrule
\multicolumn{2}{@{}l}{\textbf{CD8A -- Tumor-infiltrating lymphocytes}} \\
\addlinespace[0.25em]

Gene evidence &
UniProt: ``coreceptor for MHC class I molecule:peptide complex'';
``interacts simultaneously with the TCR and MHC class I.'' \\

Reviewed morphology assumption &
T-cell antigen-receptor / surface-signaling complexes, when expressed at
\textbf{HIGH} levels, are associated with increased
\textbf{tumor-infiltrating lymphocytes}. \\

PathGeneLink output &
\textbf{Member}; instantiated association:
\textbf{CD8A HIGH $\rightarrow$ increased TILs}. \\

Clinician evaluation &
Final clinician reference label: \textbf{Valid}. \\

\end{longtable}

\section{Evaluation-set construction and blinded review}
\label{app:evaluation-set}

\subsection{Evaluation-set construction}

The evaluation unit is a \emph{gene--assumption pair}. We evaluated 12 genes:
CDH1, MKI67, ACTA2, CD8A, EGFR, PTEN, CTNNB1, BRCA1, PECAM1, MS4A1, CD68, and
VIM.

The evaluation set is constructed from three groups:

\begin{enumerate}
    \item \textbf{PathGeneLink-emitted associations (110).}
    These are the associations emitted by the final PathGeneLink configuration
    for the 12 genes.

    \item \textbf{Similarity-selected non-emitted candidates (136).}
    These are non-emitted assumptions selected for topical similarity between
    the gene's evidence and the functional-class description of the
    assumption.

    \item \textbf{Random non-emitted candidates (136).}
    These are sampled from the remaining non-emitted candidates and provide a
    broader comparison set.
\end{enumerate}

Of these 382 candidate pairs, 379 received a definite clinician label and were
used for evaluation.

Similarity is used only to select plausible non-emitted candidates for review;
it is not a PathGeneLink score and does not use the model's membership verdict
or confidence. Candidate similarity is based on embedding representations of
the gene evidence and the assumption's gene-functional-class description.
Selection is capped across morphology features to prevent a single feature from
dominating the similarity-selected subset.

\subsection{Blinded clinician review}

A single expert clinician evaluated all 379 pairs. For each pair,
the clinician was shown the gene evidence and the reviewed assumption and judged
whether the gene belonged to the functional class described by that
assumption.

The PathGeneLink decision and the pair's sampling group
(emitted, similarity-selected, or random) were hidden during review. The
clinician therefore performed the same gene-membership judgment as the automated
system without seeing its answer. These judgments provide the reference labels
used for evaluation.

\section{Statistical analysis and matched baselines}
\label{app:statistics}

\subsection{Agreement and classification metrics}

The primary agreement statistic is Cohen's $\kappa$,

\[
\kappa = \frac{p_o-p_e}{1-p_e},
\]

where $p_o$ is observed agreement and $p_e$ is agreement expected from the
marginal label frequencies.

Precision and recall are computed with the clinician review as
the reference labeling:

\[
\mathrm{Precision} = \frac{TP}{TP+FP},
\qquad
\mathrm{Recall} = \frac{TP}{TP+FN}.
\]

For PathGeneLink, the 379-pair comparison contains 85 true positives,
25 false positives, 17 false negatives, and 252 true negatives, yielding
precision $=85/110=0.77$, recall $=85/102=0.83$, and raw agreement
$=(85+252)/379=88.9\%$.

\paragraph{Bootstrap confidence intervals.}
Confidence intervals were obtained by a nonparametric case bootstrap over the
$n=379$ gene--assumption pairs. In each of $B=10{,}000$ replicates we resampled
the 379 pairs with replacement and applied the same resampled indices to both the
model (or clinician) labels and the reference labels, so that each pair is the
resampling unit and its paired labels are kept together. Cohen's $\kappa$,
precision, and recall were recomputed on each resampled set; replicates in which a
statistic was undefined (e.g.\ $\kappa$ when $p_e$ degenerates to $1$, or precision
with no emitted positives) were discarded. The reported 95\% intervals are the
$2.5$th and $97.5$th percentiles of the resulting bootstrap distribution. The
random number generator was seeded once ($\texttt{seed}=20260830$) for
reproducibility.

\subsection{Matched LLM baselines}

All four conditions in Table~\ref{tab:baseline-comparison} are evaluated on the
same 379 gene--assumption pairs and are scored against the same
binary clinician reference labels. They differ in the information supplied to
the model:

\begin{itemize}
    \item \textbf{Direct LLM:} neither retrieved gene evidence nor the reviewed
    biological assumption is supplied.

    \item \textbf{LLM + gene evidence:} retrieved gene evidence is supplied,
    but the reviewed assumption is withheld.

    \item \textbf{LLM + reviewed assumption:} the reviewed assumption is
    supplied, but retrieved gene evidence is withheld.

    \item \textbf{PathGeneLink:} both retrieved gene evidence and the reviewed
    assumption are supplied.
\end{itemize}

The evaluation set, clinician reference labels, and scoring procedure are
held fixed across conditions.

\section{Prompts and model configuration}
\label{app:prompts}

\subsection{Assumption-generation models}

The morphology-assumption pipeline separates candidate authoring, literature
retrieval, and evidence verification. The model configuration used for the
reported resource is summarized in Table~\ref{tab:app-assumption-models}.

\begin{table}[h]
\caption{Model roles in morphology-assumption construction.}
\label{tab:app-assumption-models}
\centering
\small
\begin{tabular}{@{}lll@{}}
\toprule
\textbf{Stage} & \textbf{Role} & \textbf{Model} \\
\midrule
Author & Candidate assumption generation & \texttt{gpt-5.5} \\
Retrieve & Literature retrieval and grounding & \texttt{gpt-5.4} \\
Verify & Independent evidence assessment & \texttt{gpt-5.5} \\
\bottomrule
\end{tabular}
\end{table}

\subsection{Gene-specific linking models}

The primary gene-specific linking configuration uses
\texttt{gpt-5.4-mini} for the generation pass and \texttt{gpt-5.5} for the
adversarial verification pass. Both passes use the same gene-membership
criterion and operate only on the information supplied in the prompt.

%%%%%%%%%%%%%%%%%%%%%%%%%%%%%%%%%%%%%%%%%%%%%%%%%%%%%%%%%%%%
%%%%%%%%%%%%%%%%%%%%%%%%%%%%%%%%%%%%%%%%%%%%%%%%%%%%%%%%%%%%
\subsection{Assumption-generation prompts}
\label{app:assumption-prompts}

The morphology-assumption pipeline uses three stages: Author, Retrieve, and
Verify. The prompt templates below are reproduced from the implementation.
Placeholders enclosed in braces are substituted at run time. Several prompts
contain shared principle blocks; these blocks are reproduced separately below.

\subsubsection{Author: causal assumptions}

\paragraph{System prompt.}

\begin{Prompt}
You are an expert histopathologist and molecular biologist.

For a given morphology/pathology feature, write a GENE-AGNOSTIC set of ASSUMPTIONS: which classes of gene FUNCTION, at which EXPRESSION DIRECTION, MECHANISTICALLY produce the feature -- the drivers whose activity causes it and the suppressors whose loss causes it. These are proposed hypotheses, framed tentatively ("may contribute to"), not settled rules.

{CLAIM_MODEL}

SCOPE -- CAUSAL ONLY. List the distinct causal mechanisms that genuinely hold -- the set a careful pathologist would enumerate, complete but not inflated. Cover mechanisms that bear on the feature through DIFFERENT kinds of biology, not several variations on one. Merge classes that are the same mechanism; do NOT split one pathway into near-identical sub-steps, and do NOT pad. Author at the level of a distinct mechanism, not its internal steps: when several steps belong to ONE program and act only by serving that same program's effect on the feature, name the program as one class rather than listing each step separately. A mechanism that RESTRAINS the feature (a brake whose loss produces it) is its own class, never folded into the drivers it opposes -- a driver-high and a brake-low are DISTINCT even when they sound similar. Include a class here ONLY if changing its expression would MECHANISTICALLY change the feature; if a class merely co-varies with the feature -- its expression tracks or predicts the feature without a mechanism that produces it -- leave it OUT here (it is a marker, authored in the separate correlative step). A useful test: if the only honest support you could ever cite is that the class's expression is ASSOCIATED WITH, PREDICTS, or IS RAISED/LOWERED IN the same cells where the feature occurs -- rather than a mechanism showing that its expression change PRODUCES the feature -- then it co-varies and belongs in the correlative step, not here. Do not upgrade an association into a cause because the class feels central or important. When unsure whether a class causes the feature or only correlates with it, do not force it into the causal set. Every claim here is relation=causal (and specificity=specific by construction).

DO NOT cite anything. Do not attach PMIDs, quotes, or a strength rating -- a later stage grounds and grades these. Your ONLY job is to enumerate the real mechanisms at full recall and reason each one clearly. You may use web search to jog your memory of the biology, but you are not collecting citations.

{DIRECTION}
If the feature term has more than one histologic meaning, state in `definition` which sense these claims address and write to that sense.

{ORGAN_SCOPE}

{GO_TAGGING}

{HONESTY}

Emit as many genuinely-distinct causal classes as are real; do not invent mechanisms to hit a number.

ANCHOR (level + style ONLY -- an unrelated feature, shown so you see the granularity and which fields to fill, NOT to be copied or pattern-matched to the real feature): feature "cellular swelling" -> a causal claim like "genes that maintain plasma-membrane ion pumps and cell-volume / osmotic homeostasis, expression_low -> ions and water are no longer extruded and the cell swells", organs ["kidney","liver"], pan_organ true, reasoning giving the mechanism.
\end{Prompt}

\paragraph{User message.}

\begin{Prompt}
Morphology feature: {feature}
{options}
\end{Prompt}

\subsubsection{Author: correlative assumptions}

\paragraph{System prompt.}

\begin{Prompt}
You are an expert histopathologist and molecular biologist.

The CAUSAL classes for this feature are already authored and given below. Author the CORRELATIVE (marker) claims: gene-function classes whose EXPRESSION MARKS/tracks the feature ITSELF -- the cell-state the feature directly IS -- WITHOUT causing it. Framed tentatively ("tends to mark").

{CLAIM_MODEL}

THREE HARD RULES:
1. DISTINCT from the causal classes below -- do not restate, rename, or lightly reword any of them. If a class mechanistically CAUSES the feature it is causal (already done); skip it here.
2. READ-OUT TEST -- tag every marker specificity = "specific" or "non_specific":
   - specific: THIS feature IS the single cell-state this marker reads out (a proliferation antigen vs. "frequent mitoses"). A lineage/histotype marker reads out LINEAGE, not the feature -- NOT specific even when the feature is common in that lineage.
   - non_specific: the feature is a SEPARATE property that merely co-varies with the marker's state in the same tumors, so ANY marker of that state would show the same correlation. KEEP these, tag honestly.
3. Markers must be REAL and genuinely citable (a later stage grounds them). An EMPTY list is the correct answer when a feature has no genuine marker; do not invent or pad.

DO NOT cite anything (no PMIDs, quotes, or strength rating) -- a later stage grounds and grades these.

DIRECTION. expression_high when high marker expression tracks the feature; expression_low when loss/low of a marker tracks it. Single direction only when the other is not real.

Use the SAME shape rule as the causal claims, from the grading options: if the low end just means the feature is ABSENT, tag every marker's `feature_pole` EMPTY and give markers of the present state only -- a marker of viable / normal / feature-absent tissue is NOT a marker of the feature. Only when BOTH ends are real opposite states may a marker read out either end (tag 'high' / 'low'). (`feature_pole` = which end of the feature, separate from `direction`, the marker gene's own expression level.)

{ORGAN_SCOPE}

{GO_TAGGING}

{HONESTY}

ANCHOR (level + style ONLY -- an unrelated feature, shown for the read-out-test distinction, NOT to be copied or pattern-matched to the real feature): feature "melanocytic differentiation" -> specific marker "melanocyte-lineage pigment-synthesis and identity genes, expression_high" (the marker reads out the melanocytic cell-state the feature IS); non_specific marker "proliferation antigen, expression_high" (co-varies via aggressiveness, not specific to this feature; tag non_specific).
\end{Prompt}

\paragraph{User message.}

\begin{Prompt}
Morphology feature: {feature}
{options}

The CAUSAL claims already authored for this feature (do NOT restate any of these; add only conceptually-distinct MARKER claims):
{causal_context}
\end{Prompt}

\subsubsection{Retrieve}

\paragraph{System prompt.}

\begin{Prompt}
You are an expert biomedical evidence retriever. You are given ONE morphology assumption authored by a pathologist. Your job is to GROUND it in the published literature, and to refine it to match what the evidence actually shows.

REFINE, DON'T DRIFT. Update the claim wherever your evidence genuinely warrants it -- reword `gene_function`/`reasoning` for accuracy, and treat `organs`/`pan_organ` as FIRST-CLASS. Leave the GO tags (`go_slim_terms`/`mf_slim_terms`/`cc_slim_terms`) EXACTLY as given -- copy them through verbatim; they were chosen from a controlled vocabulary you do not have here, so do not invent or change them. But every other change must be EVIDENCE-DRIVEN and JUSTIFIED: if you change anything, `change_reason` must give the reason for the edit, grounded in the evidence you found (you need NOT list what changed -- that is detected automatically; give only the WHY). Do NOT change the claim for style, and do NOT narrow the gene-function class more than the evidence requires (a needlessly narrow class matches fewer genes downstream). If the evidence fits the claim as authored, leave it and leave `change_reason` empty.

ORGAN SCOPE IS DECIDED BY THE BIOLOGY, NOT BY WHICH PAPER YOU HAPPENED TO FIND. The organ where a study was done is NOT automatically the organ where the mechanism operates: finding the link demonstrated in one organ does NOT, on its own, make the mechanism organ-specific. Ask whether the mechanism itself depends on tissue-specific machinery (a lineage program, hormone axis, secretory product -> genuinely organ-scoped) or on machinery essentially all tissues share (-> broad, likely pan-organ), and set the scope to THAT -- see ORGAN SCOPE below and the search rule for confirming span. Narrowing a genuinely broad mechanism down to the single organ of the one paper you found is as much an error as over-claiming breadth: it makes the rule silently miss every other tissue where it holds.

{CLAIM_MODEL}

SEARCH HARD, ONCE. Use web search THOROUGHLY in this single pass -- aim for at least 8-12 distinct searches, and do not stop early: the most common failure is under-searching and returning too few sources, which loses real support. Vary the angle every time: the exact feature, its synonyms, the gene-class and its member genes, organ-specific instantiations, and -- when one abstract only reaches part of the link -- search SEPARATELY for the missing piece (the expression-direction effect, an intermediate state, or the feature itself). Critically, hunt for a paper that ties the EXPRESSION DIRECTION (high/low) of this gene-class to the feature -- not just the pathway in general. There is no second retrieval round, so exhaust the angles now; only stop when repeated new searches stop yielding anything on-target (not merely when you have one source). Every PMID you return must be one you actually found via search for THAT paper -- never write a PMID from memory (a wrong number resolves to the wrong paper and is discarded).

TEST THE ORGAN SPAN, DON'T ASSUME IT. Once you have the link in one organ, spend one or two searches checking whether it also holds in OTHER, unrelated tissues -- search the mechanism (not the organ you already have) paired with a different site, or look for a review stating the mechanism generally. If you confirm it in >=2 unrelated organs, list them and set pan_organ per the ORGAN SCOPE rule; if a review establishes it broadly, use the pan-tissue sentinel. If you genuinely find it in only one organ despite looking, keep the scope narrow -- but then `change_reason` must say you checked other organs and did not find them, so a single-organ scope reflects a real search, not an unchecked assumption.

{EVIDENCE_BAR}

RETURN:
- `core`: the claim's identity, refined to match the evidence (reworded / re-scoped organs as warranted), otherwise as given; GO tags copied through verbatim.
- `evidence`: the abstracts that support it, each with a verbatim `quote` from that abstract and a `hop` tag ("full" / "function" / "intermediate" / "feature"). Return an EMPTY evidence list if after a genuine hard search nothing meets the bar. "Nothing found" is an honest, wanted answer; do not attach a weak paper to avoid an empty list.
- `change_reason`: "" if you changed nothing; else the evidence-grounded WHY for your edit (organ/scope changes especially). Give only the reason; WHAT changed is detected automatically.

Do NOT assign any strength/tier rating -- an independent stage grades the evidence you return.

{ORGAN_SCOPE}
\end{Prompt}

\paragraph{User message.}

\begin{Prompt}
Morphology feature: {claim.feature}

THE CLAIM to ground:
- gene_function: {claim.gene_function}
- direction: {claim.direction}
- relation: {claim.relation}
- reasoning: {claim.reasoning}
- organs: {claim.organs} | pan_organ: {claim.pan_organ}
- go_slim_terms: {claim.go_slim_terms} | mf_slim_terms: {claim.mf_slim_terms} | cc_slim_terms: {claim.cc_slim_terms}
\end{Prompt}

If the initial retrieval pass returns no surviving evidence, the following
retry instruction is appended to the user message:

\begin{Prompt}
RETRY: a first thorough search returned no abstract that meets the evidence bar. Do NOT give up yet -- search again with DIFFERENT angles you have not tried: the specific member genes of this class, the exact effect at this expression direction, an intermediate state, close synonyms of the feature, and organ-specific instantiations. Only return an empty set if this second, genuinely different search also finds nothing on-target.
\end{Prompt}

\subsubsection{Verify}

\paragraph{System prompt.}

\begin{Prompt}
You are a careful independent biomedical fact-checker. You are given a morphology assumption (a gene-function class, at an expression direction, linked to a feature) and a SET of published abstracts retrieved as evidence. You did NOT author the claim; ignore how confident it sounds and judge ONLY what the evidence shows. Every quote is already verified to appear in its abstract, so trust the quotes are real; your job is to decide how well they support the claim.

{EVIDENCE_BAR}

THE CORE QUESTION (ask this of the evidence, per abstract, gene-agnostic): does an abstract back the claim's mechanism-class AND direction for THIS feature? A paper need NOT name a specific gene -- establishing the function/pathway -> feature link at the class level is enough. A SINGLE abstract that states the link is sufficient to support the claim; you do NOT need multiple papers, and you do NOT need to decompose the claim into sub-steps and evidence each one separately.

FEATURE MATCH -- the evidence must actually reach THIS SPECIFIC feature (as named, and as scoped in any feature-scope note), not merely a RELATED, BROADER, or ADJACENT process. If an abstract supports the mechanism but the morphology/endpoint it reaches is a DIFFERENT feature, or only a looser proxy for this one, it does NOT support this claim -- do not stretch it. (E.g. evidence about a general process is not evidence for a specific structural feature that merely tends to accompany it.) When the evidence is only about an adjacent feature, grade it weak/ungrounded and say which feature it actually supports.

ASSIGN one tier:
- established: a single abstract (or clearly the field's textbook knowledge reflected in the evidence) directly states the function@direction -> feature link -- a direct, well-established link.
- supported: the evidence backs the link, but it holds via an intermediate step or only in certain contexts/organs, OR no single abstract states the whole link yet two abstracts together clearly close it through the SAME named intermediate. A single supporting abstract normally lands here (or established if the link is direct/textbook).
- weak: the evidence is only topically related or partial -- it discusses the mechanism OR the feature but does not actually establish this function@direction -> feature link; or a claimed two-paper chain does not clearly connect through the same intermediate.
- ungrounded: no abstract supports the link, OR the only evidence is at the wrong lever (a mutation/deletion/knockout rather than an EXPRESSION-level change). This means "unsupported by the evidence retrieved" -- NOT proof the biology is absent (a search miss reads the same), so do not phrase it as a biological gap.

DEFAULT TO SINGLE-PAPER JUDGMENT. Most claims are supported by ONE good abstract -- grade those on that abstract alone; do not withhold 'supported' merely because you can imagine intermediate steps that aren't separately cited. Chaining two papers is a FALLBACK for when no single abstract suffices -- never a requirement, and it must never DOWNGRADE a claim that a single abstract already supports.

CHAIN (fallback only): if -- and only if -- no single abstract states the whole link but two or more abstracts together close it, name the shared intermediate in `intermediate` (a deterministic check confirms it appears in the joined quotes; if not, the tier is capped to weak). Leave `intermediate` EMPTY when a single abstract supports the claim (the common case).

PRUNE: put in `drop_pmids` the PMIDs of evidence items that are off-topic or name-drops on your independent read. Leave it empty to keep all. Refer to items by PMID only -- do NOT re-transcribe quotes.

Do NOT reassess organ scope -- that was decided during retrieval; judge only the evidence's strength.

{HONESTY}

In `verdict_note`, state briefly what supports the claim (which abstract backs the link) or, for weak/ungrounded, what is missing (topical-only, wrong-lever, or a chain that doesn't connect).
\end{Prompt}

\paragraph{User message.}

\begin{Prompt}
THE CLAIM:
- feature: {claim.feature}
- gene_function: {claim.gene_function}
- direction: {claim.direction}
- relation: {claim.relation}
- reasoning: {claim.reasoning}
- organs: {claim.organs} | pan_organ: {claim.pan_organ}

THE EVIDENCE (each quote is already verified present in its abstract):
{ev_lines}

To prune a non-supporting item, put its PMID in `drop_pmids` (do NOT re-transcribe quotes).
\end{Prompt}

\subsubsection{Shared assumption-generation prompt blocks}

\paragraph{\texttt{CLAIM\_MODEL}.}

\begin{Prompt}
A CLAIM links a gene-FUNCTION class to a morphology feature at an expression DIRECTION: "[a kind of gene], when expressed [high/low], [produces / tracks] [the feature]".
- gene-agnostic: it names a CLASS of gene by what it does ("genes that build cell-cell adhesion junctions"), matching ANY gene with that role -- never a single named gene (a specific gene may appear only as an illustration inside the prose).
- gene-agnostic is NOT organ-agnostic: "androgen-responsive stromal-support genes" matches any such gene yet applies only to certain organs. The organ scope is a separate axis (see ORGAN SCOPE).
This predicts a tissue's morphology from its gene EXPRESSION (not mutations). Reason from the biology in neoplastic OR normal tissue -- do not assume malignancy.
\end{Prompt}

\paragraph{\texttt{DIRECTION}.}

\begin{Prompt}
Expression is directional and role-dependent, and the two directions are INDEPENDENT routes to the same feature -- a feature can arise from HIGH expression of a gene that DRIVES it AND, separately, from LOW expression of a different gene that SUPPRESSES/removes it. Many features are balance-controlled (a production-vs-removal or build-vs-degrade steady state), so BOTH arms are genuinely real and you should author BOTH as distinct claims:
  - an accumulation/deposition feature: HIGH synthesis/uptake OR LOW clearance/degradation;
  - a loss-of-structure feature: LOW of what builds/maintains it OR HIGH of what degrades it.
Author only ONE direction when the opposite arm has NO genuine mechanism -- i.e. losing the opposite gene gives a different feature, not this one (e.g. a synthesis-dominated feature like pigment or mucin accumulation has no distinct "clearance gene" whose loss independently produces it; a proliferation read-out is intrinsically unidirectional). Do not manufacture a symmetric opposite that has no real biology.
READ THE GRADING OPTIONS to decide how to author (they are the mutually-exclusive levels the feature is scored on):
  - If the lowest option just means the feature is ABSENT (it says "absent", or the low end is "none" / nothing there), the feature is present-or-not: author ONLY the mechanisms that PRODUCE the present state, and leave `feature_pole` EMPTY.
  - If BOTH ends are real opposite STATES a tissue can be in (neither is mere absence), the feature is a two-ended spectrum: author each end's OWN producing biology -- not the negation of the other end -- and tag `feature_pole` "high" (the greater / more-atypical end) or "low" (the lesser / more-ordered end).
If no options are given, author the feature as present-or-not (feature_pole empty) unless it is plainly a two-ended spectrum. (`feature_pole` = which end of the feature; it is separate from `direction`, the gene's own expression level -- an ordered-state gene expressed high can produce the "low" end.)
\end{Prompt}

\paragraph{\texttt{EVIDENCE\_BAR}.}

\begin{Prompt}
WHAT COUNTS AS SUPPORT. A paper supports a claim only if, reading its ABSTRACT (title + abstract text, not the full body, not your own inference), the link "[gene-function class] expressed [high/low] -> [this feature]" is actually stated or directly shown. It need not use our exact words, but a reader must see the connection in the abstract itself.
NOT support (be strict -- these are the common near-misses):
  - a NAME-DROP: the abstract mentions the gene class OR the feature but does not connect them;
  - MECHANISM-ONLY: shows the pathway/expression change but never reaches this morphology feature;
  - FEATURE-ONLY: shows the morphology but not the gene-expression cause/correlate;
  - WRONG LEVER: shows a MUTATION / deletion / knockout / LOH, not an EXPRESSION-level change;
  - CORRELATION vs CAUSATION mismatch: a stated correlation supports a CORRELATIVE claim, but does NOT by itself support a CAUSAL claim.
A claim may be supported by a SET of abstracts that compose the chain end-to-end (one for the mechanism/expression half, one linking the intermediate to the feature) -- tag each with its hop. This is preferred over forcing a single paper to say everything.
\end{Prompt}

\paragraph{\texttt{ORGAN\_SCOPE}.}

\begin{Prompt}
ORGAN SCOPE. A class is gene-agnostic but often organ-SCOPED. Judge, from the biology: does producing the feature rest on machinery essentially all tissues have (adhesion, apoptosis, mitosis) -> leans pan-organ; or does it REQUIRE a tissue-specific factor (a lineage program, hormone receptor, secretory product) -> organ-specific? Keep a general mechanism separate from an organ-specific instantiation of it (loss of trophic support -> atrophy is general; its androgen-driven form is prostate-only).
NO LEAPS: set pan_organ=true ONLY when grounded broadly -- evidence in >=2 unrelated organs (list them in `organs`) OR a review / general-pathology source. Do NOT infer pan-organ from a single-organ paper. If your only basis is one organ, put it in `organs` and leave pan_organ=false.
Name `organs` with STANDARD organ / tissue names in plain English (e.g. "colon", "breast", "lung", "prostate") -- one per organ; write them the way a pathologist would say them, there is no fixed list to match. Use '{PAN_TISSUE}' as the sole entry only when pan_organ is true and no specific organ list applies.
\end{Prompt}

\paragraph{\texttt{HONESTY}.}

\begin{Prompt}
State relationships precisely, not deterministically. Spell out the chain including any intermediate step in contributory wording ("low expression of a membrane ion-pump gene lets ions and water accumulate; IF osmotic balance is lost, this contributes to cell swelling"), not "low expression -> swollen cell". Use "contributes to / can favor / tends to mark", not "causes", when a step or context is required. If a class supports only PART of a broad feature, say so. Do not inflate.
\end{Prompt}

\paragraph{\texttt{GO\_TAGGING}.}

\begin{Prompt}
Tag each claim against the GO-slim vocabulary in THREE INDEPENDENT aspects (exact NAMEs, verbatim; typically 1-3 each, up to 4 if genuinely multi-functional). Favor RECALL -- include a term if it honestly applies -- but never pad, and leave an aspect EMPTY when nothing fits. GO never links the aspects; tag each on its own.
  go_slim_terms = Biological Process (the broader process this function serves)
  mf_slim_terms = Molecular Function (what it does biochemically)
  cc_slim_terms = Cellular Component (where in the cell it acts; often empty -- fine)

BIOLOGICAL PROCESS:
{GO_BP_LINES}

MOLECULAR FUNCTION:
{GO_MF_LINES}

CELLULAR COMPONENT:
{GO_CC_LINES}
\end{Prompt}

%%%%%%%%%%%%%%%%%%%%%%%%%%%%%%%%%%%%%%%%%%%%%%%%%%%%%%%%%%%%
\subsection{Gene-specific linking prompts}
\label{app:linking-prompts}

The gene-specific linking stage makes one substantive judgment: whether a
specific gene belongs to the functional class defined by a candidate
assumption, using only the supplied evidence for that gene. The
function-to-morphology relationship and expression direction are inherited from
the clinician-reviewed assumption and are not re-derived during this stage.

\subsubsection{Generation pass}

\paragraph{System prompt.}

\begin{Prompt}
You are an expert molecular pathologist. You are given ONE gene and ONE tissue morphology FEATURE, plus a set of candidate ASSUMPTIONS about that feature. Each assumption is a gene-agnostic, literature-reviewed hypothesis: 'a gene of function-class F, at expression STATE (high or low), drives this feature toward a specific TARGET' (the target names the exact end, e.g. 'poorer tumor differentiation' vs 'better tumor differentiation'). The class->feature link -- and its direction -- is an ALREADY-REVIEWED, accepted assumption; you do NOT re-derive it, judge its direction, or rate a confidence. Your task is narrower and is the ONLY thing you do: for THIS gene, decide whether it is a member of each assumption's gene-function class, judged solely from the gene's own database evidence. Invent no outside facts.

{MODALITY_LINE}

{GENE_EVIDENCE_NOTE}
{CONTRACT}

{EXAMPLE}
\end{Prompt}

\subsubsection{Adversarial verification pass}

\paragraph{System prompt.}

\begin{Prompt}
You are an adversarial reviewer RE-CHECKING a first-pass set of membership verdicts. You are given the ORIGINAL, identical inputs the first pass saw (gene, feature, the gene's database evidence with ids, and the candidate assumptions) plus the first pass's 'member' verdicts. Judge against the EXACT SAME contract below, never a private one. For EACH candidate listed, try to REFUTE the 'member' verdict: independently re-name the class criterion and the gene's cited activity, then test one against the other. KEEP 'member' only if the gene's OWN evidence genuinely shows the class's DEFINING activity in the stated direction; otherwise downgrade the verdict to broad_overlap, incidental, opposite_direction, or unrelated as appropriate. Return one verdict per listed candidate (echoing its short id).

{MODALITY_LINE}

{GENE_EVIDENCE_NOTE}
Spend your effort where it matters most: is the gene GENUINELY a directional member, grounded in its OWN evidence? (Citation presence is enforced deterministically downstream -- do NOT re-check quotes character-by-character.) Ground everything ONLY in the supplied inputs; add no outside facts; never assert membership from your own knowledge. A sound 'member' should pass through unchanged.

===== THE CONTRACT (identical to the first pass) =====
{CONTRACT}

{EXAMPLE}
\end{Prompt}

\subsubsection{Shared membership contract}

\begin{Prompt}
TASK: for EACH candidate assumption you are given, decide whether THIS gene is a member of the assumption's gene-function class. Return one verdict per candidate, in the order shown. This is the ONLY judgment you make -- you do NOT judge the class->feature direction, a relation type, or a confidence.

Work through the four fields per candidate TOP-TO-BOTTOM, so the verdict is the conclusion of the reasoning above it (never a snap choice rationalized afterward):

1) class_criterion -- BEFORE looking at the gene, state the single DEFINING activity the class names, together with its DIRECTION. A class names an activity with a direction (schematically 'genes that BUILD structure S', or 'genes that ENFORCE process P'). Name what a gene must do to belong -- and what would NOT qualify: merely participating in a broad pathway the class sits in, showing the activity only in passing / as one of several roles, or performing the OPPOSITE activity. Commit to this bar first, gene-blind.

2) membership_citations -- now read THIS gene's own database evidence and quote the line(s) that bear on that criterion. Cite by evidence id; quote verbatim; never paraphrase or invent. If the gene's evidence contains nothing that bears on the criterion, leave this empty. Your only citation job is the gene's membership -- the class->feature half is an already-reviewed, accepted assumption you never cite.

3) membership_reasoning -- ONE sentence: measured against the criterion in (1), does the cited evidence show the class's DEFINING activity in the class's stated DIRECTION, or does it fall short -- a broad-pathway/family overlap, an incidental/secondary role, the OPPOSITE direction, or unrelated? Judge ONLY from the supplied evidence, never from training knowledge or how 'canonical' the gene is.

4) verdict -- the conclusion of (3), exactly one of:
     member             -- the gene's OWN evidence shows the class's defining activity in the stated direction (grounded by >=1 citation).
     broad_overlap      -- the gene only participates in a broad pathway/family the class belongs to, not the defining activity itself.
     incidental         -- the gene shows the activity only in passing / as a secondary or hedged role, not as a defining function.
     opposite_direction -- the gene acts on the same process but in the OPPOSITE direction (it belongs to a different class).
     unrelated          -- the gene's evidence does not bear on the class activity at all.

Only 'member' is an accepted association; the other four are grounded rejections. Membership is EXACT, not approximate, and evidence-only: if you cannot cite a supplied quote that meets the criterion, the verdict is NOT 'member'. Most candidates are non-members -- a well-grounded rejection is as valuable as a link; do not stretch to 'member' to avoid a rejection, and never base a call on how familiar the gene is.
\end{Prompt}

\subsubsection{Shared few-shot example}

\begin{Prompt}
EXAMPLES -- illustrative SHAPE only. The biology is abstract placeholders (structure S, process P,
cell-state Q). Do NOT map them to any real gene or feature, and do NOT treat 'member' as the expected answer
(most candidates are non-members).

Class in A-D: "genes that BUILD structure S", state=expression_low.

A) class_criterion: a member must ASSEMBLE / MAINTAIN structure S; degrading S, or a broad role in the
   compartment S lives in, does not qualify.
   evidence [DB - FUNCTION]: "assembles and maintains structure S."
   membership_reasoning: the evidence states the defining build-S activity directly.
   verdict: member.   (membership_citations: the "assembles ... S" quote.)

B) evidence [DB - FUNCTION]: "a negative regulator that DISASSEMBLES structure S."
   membership_reasoning: acts on S but in the OPPOSITE direction -- a disassembler is not a member of a build-S
   class.
   verdict: opposite_direction.

C) evidence [DB - FUNCTION]: "participates in compartment-P homeostasis, of which structure S is one part."
   membership_reasoning: only a broad-pathway/compartment role, not the defining build-S activity.
   verdict: broad_overlap.

D) evidence [DB - FUNCTION]: "primarily performs activity Z; also reported, in passing, to stabilize S in one
   context."
   membership_reasoning: the build-S activity is a secondary / in-passing role, not the gene's defining function.
   verdict: incidental.

E) Class "genes MARKING cell-state Q", state=expression_high. evidence describes only an unrelated activity (a
   product of some process Z).
   membership_reasoning: nothing in the evidence bears on marking cell-state Q.
   verdict: unrelated.
\end{Prompt}

\subsubsection{Modality line}

\begin{Prompt}
MODALITY: this gene's expression is measured as {assay}; 'expression_high/low' refers to its {state_noun}.
\end{Prompt}

\subsubsection{Gene-evidence reading instruction}

\begin{Prompt}
Each gene-evidence line is labeled [SOURCE - section]: the database and the EXACT section the quote came from (e.g. 'UniProt - FUNCTION' says what the gene DOES; 'HPA - RNA tissue specificity' or 'HPA - Subcellular location' says WHERE / at what level / in which cell types it sits). Read each quote in its proper context from that section name -- a tissue/location section says WHERE the gene sits, a FUNCTION section says what it DOES; do not mistake one for the other.
Some lines end with an automated tag in angle brackets, e.g. <function>, <expression: high -- colon>, or <morphology>. These come from a rough automated pre-scan and may be wrong or incomplete; use a tag only as a pointer to save scanning. Base every judgment on the quoted text itself and your own reasoning, and ignore any tag the quote does not support.
\end{Prompt}

\subsubsection{User message}

The same base user context is provided to the generation and verification
passes.

\begin{Prompt}
GENE: {gene}{marker_suffix}
MORPHOLOGY FEATURE: {feature}
Feature definition: {feature_def}

=== THIS GENE'S DATABASE EVIDENCE (the ONLY basis for membership; cite items by their id) ===
{gene_evidence}

=== CANDIDATE ASSUMPTIONS FOR THIS FEATURE (return ONE verdict per candidate, echoing its short id, e.g. A3) ===
{candidates}
\end{Prompt}

Each gene-evidence item is rendered as:

\begin{Prompt}
  - [{src_label}] "{quote}"  (cite as {evidence_id}){pmid_suffix}{annotation}
\end{Prompt}

When no gene evidence is available:

\begin{Prompt}
(no database evidence supplied for this gene -- membership cannot be grounded, so no candidate can be a member.)
\end{Prompt}

Each candidate assumption is rendered as:

\begin{Prompt}
ASSUMPTION id={prompt_id}
  {gene_class}, when expressed at {LEVEL} levels, may contribute to {target}.
\end{Prompt}

For the adversarial verification pass, the following section is appended:

\begin{Prompt}
=== FIRST-PASS 'member' VERDICTS (re-check each; keep 'member' only if it genuinely holds, else downgrade) ===
{members_json}
\end{Prompt}

%%%%%%%%%%%%%%%%%%%%%%%%%%%%%%%%%%%%%%%%%%%%%%%%%%%%%%%%%%%%
\subsection{Matched-baseline prompts}
\label{app:baseline-prompts}

All three matched baselines are evaluated on the same 379 gene--assumption
pairs and return the same structured prediction: a one-sentence
\texttt{reasoning} and a binary \texttt{verdict} in
$\{\texttt{associated},\texttt{not\_associated}\}$, where
\texttt{associated} is the positive prediction. The conditions differ only in
which information is supplied to the model.

The common per-pair claim has the form:

\begin{Prompt}
When {gene} is expressed at {HIGH/LOW} {assay} levels, it contributes to {morphology target}.
\end{Prompt}

The shared modality line is the same as that used for gene-specific linking:

\begin{Prompt}
MODALITY: this gene's expression is measured as {assay}; 'expression_high/low' refers to its {state_noun}.
\end{Prompt}

\subsubsection{Direct LLM}

This condition receives neither retrieved gene evidence nor the reviewed
assumption.

\paragraph{System prompt.}

\begin{Prompt}
You are an expert molecular pathologist. Using ONLY your own expert knowledge -- you are given no database evidence and no external documents -- judge whether the stated association is valid: is the gene, at the stated expression level, a genuine contributor to the stated morphological feature in tumor tissue? Answer 'associated' or 'not_associated'. Do not assume 'associated' is the expected answer; many plausible-sounding associations are not real.

{MODALITY_LINE}
\end{Prompt}

\subsubsection{LLM + gene evidence}

This condition receives the gene's database evidence but does not receive the
reviewed biological assumption. The same gene-evidence reading instruction used
by PathGeneLink is included.

\paragraph{System prompt.}

\begin{Prompt}
You are an expert molecular pathologist. Judge whether the stated association is valid, grounding your decision in the supplied gene DATABASE EVIDENCE plus sound reasoning -- not in memorized canonical associations. Is the gene, at the stated expression level, a genuine contributor to the stated morphological feature in tumor tissue? Answer 'associated' or 'not_associated'. Do not assume 'associated' is the expected answer; base the call on what the evidence actually supports.

{MODALITY_LINE}

{GENE_EVIDENCE_NOTE}
\end{Prompt}

\subsubsection{LLM + reviewed assumption}

This condition receives the clinician-reviewed, gene-agnostic assumption but
does not receive the gene database evidence. The assumption's supporting
literature is withheld.

\paragraph{System prompt.}

\begin{Prompt}
You are an expert molecular pathologist. You are given a general, clinician-reviewed hypothesis of the form 'genes of a function-class, at a stated expression level, contribute to a morphology outcome', plus ONE gene. Using ONLY your own expert knowledge of the gene -- you are given no database evidence -- judge whether THIS gene belongs to that function class, and therefore whether the stated association holds for it. Answer 'associated' or 'not_associated'. Do not assume 'associated' is the expected answer; a gene often does not belong to the class.

{MODALITY_LINE}
\end{Prompt}

\subsubsection{Baseline user inputs}

All baseline conditions receive a common header containing the gene, morphology
feature, and feature definition. The Direct LLM condition receives the
gene-specific association to judge. The LLM + gene evidence condition receives
the same association together with the gene evidence. The LLM + reviewed
assumption condition instead receives the reviewed gene-agnostic assumption and
asks whether the specified gene belongs to its functional class.

For the gene-evidence condition, the evidence is presented under:

\begin{Prompt}
THIS GENE'S DATABASE EVIDENCE (the ONLY basis for your judgment)
\end{Prompt}

The full PathGeneLink condition differs from these matched baselines in that it
receives both the source-grounded gene evidence and the clinician-reviewed
functional-class assumption.

%%%%%%%%%%%%%%%%%%%%%%%%%%%%%%%%%%%%%%%%%%%%%%%%%%%%%%%%%%%%

% \newpage
% \input{checklist.tex}

\end{document}